\pdfoutput=1

\documentclass[11pt]{article}

\usepackage[final]{acl}

\usepackage{times}
\usepackage{latexsym}

\usepackage[T1]{fontenc}

\usepackage[utf8]{inputenc}

\usepackage{microtype}

\usepackage{inconsolata}

\usepackage{graphicx}

\usepackage{wrapfig,lipsum,booktabs}
\usepackage{amsmath}
\usepackage{authblk} 
\usepackage{graphicx}
\usepackage{amsmath, amsfonts, amssymb, amsthm}
\usepackage{multirow}
\usepackage{makecell}
\usepackage{algorithm}
\usepackage[noend]{algpseudocode}
\algnewcommand{\LeftComment}[1]{\Statex \(\triangleright\) #1}
    \algnewcommand\algorithmicto{\textbf{to}}
    \algnewcommand\To{\algorithmicto{} }
    \algnewcommand\algorithmicswitch{\textbf{switch}}
    \algnewcommand\algorithmiccase{\textbf{case}}
    \algdef{SE}[SWITCH]{Switch}{EndSwitch}[1]{\algorithmicswitch\ #1\ \algorithmicdo}{\algorithmicend\ \algorithmicswitch}%
    \algdef{SE}[CASE]{Case}{EndCase}[1]{\algorithmiccase\ #1}{\algorithmicend\ \algorithmiccase}%
    \algtext*{EndSwitch}%
    \algtext*{EndCase}%

\usepackage{algorithmicx} 
\algnewcommand\NameSty[1]{\textproc{#1}}

\usepackage[T1]{fontenc}
\usepackage[utf8]{inputenc}
\usepackage{microtype}
\usepackage{inconsolata}

\usepackage{soul}
\definecolor{lightblue}{rgb}{0, 1, 1}
\definecolor{lightpink}{rgb}{1, 0.6, 1}
\definecolor{lightgreen}{rgb}{0, 1, 0.4}
\definecolor{lightorange}{rgb}{1, .75, 0}

\usepackage{wrapfig}

\title{T\scalebox{0.85}{HREAD}: Thinking Deeper with Recursive Spawning}

\author{First Author \\
  Affiliation / Address line 1 \\
  Affiliation / Address line 2 \\
  Affiliation / Address line 3 \\
  \texttt{email@domain} \\\And
  Second Author \\
  Affiliation / Address line 1 \\
  Affiliation / Address line 2 \\
  Affiliation / Address line 3 \\
  \texttt{email@domain} \\}

\author{
 \textbf{Philip Schroeder},
 \textbf{Nathaniel Morgan},
 \textbf{Hongyin Luo},
 \textbf{James Glass}
\\
\\
 MIT Computer Science and Artificial Intelligence Lab, Cambridge MA, USA
}

\begin{document}
\maketitle
\begin{abstract}
Large language models (LLMs) have shown impressive capabilities across diverse settings, but still struggle as the length and complexity of the context increases. To address this challenge, we propose Thinking Recursively and Dynamically (ThReaD). \NameSty{Thread} frames model generation as a thread of execution that, based on the context, can run to completion or dynamically spawn new threads. By spawning, threads can offload work (e.g., thinking, retrieving information) to child threads, which only return tokens needed for the parent thread to do its work.  
We apply \NameSty{Thread} in the settings of LLM task solving and question answering, where the dynamic threading allows the model to recursively decompose the given task or question into progressively simpler sub-problems that can be solved by separate child threads. We test \NameSty{Thread}, implemented using a few-shot learning approach, on diverse benchmarks for agent tasks and data-grounded question answering. \NameSty{Thread} achieves state-of-the-art performance with GPT-4 and GPT-3.5 on these benchmarks, including ALFWorld, TextCraft, and WebShop, along with two new benchmarks, DataCommons QA and MIMIC-III ICU QA. In addition, \NameSty{Thread} outperforms existing frameworks by 10\% to 50\% absolute points with smaller models, including Llama-3-8b and CodeLlama-7b. 
\end{abstract}

\section{Introduction}

\renewcommand{\thefootnote}{}  
\footnotetext{Correspondence to Philip Schroeder at \href{mailto:pschro@mit.edu}{pschro@mit.edu}. Source code is available at \url{https://github.com/philipmit/thread}.}
\renewcommand{\thefootnote}{\arabic{footnote}}  

Large Language Models (LLMs) have shown success in diverse settings \cite{wei2022chain, huang2023towards}, but their performance degrades as context length and complexity grows \cite{dziri2023faith, liu2024lost, qin2023nlp}. This constraint limits their efficacy in settings that require more work (thinking, retrieving information, analyzing, etc.) than can fit into a concise line of generation. To address this limitation, we propose Thinking Recursively and Dynamically (ThReaD).

\NameSty{Thread} is a general framework where model generation is treated as a thread of execution that, based on the context, can independently run to completion or dynamically spawn new threads in a recursive fashion. When a thread spawns a child, the child generates conditioning on context that derives from the parent's token sequence. Spawning child threads allows work, such as internal thinking or interacting with an external environment, to be completed on behalf of the parent, without directly adding to the parent's context. Child threads return only the information (tokens) needed for the parent to complete its work. In effect, spawning enables the model to dynamically adapt the amount of work or intermediate computational steps used to produce different parts of its token sequence.

The synchronization and spawning mechanisms of \NameSty{Thread} can vary based on the setting. Figure 1 shows an example of applying \NameSty{Thread} in a synchronous setting, where a parent waits for a child in a form analogous to \texttt{Thread.join()} in multithreaded programming. In this example, a parent thread pauses generation until the child execution completes and the child returns output tokens that are appended directly to the token sequence of the parent before the parent proceeds generation.
\begin{figure*}
  \centering
  \includegraphics[width=1\textwidth]{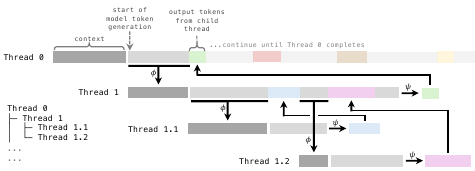}
  \caption{\textbf{\NameSty{Thread} with join synchronization.} \NameSty{Thread} frames model generation as an execution thread that can dynamically spawn new threads. In the example with join synchronization, when a thread spawns a child, it pauses generation until feedback is returned. Child threads generate starting from context derived from their parent's token sequence. When the child completes, it returns output tokens (colored bars), which are added to the context of the parent before it continues generating. $\phi$ and $\psi$ are  functions that control information flow from parent to child and from child to parent, respectively, by defining the tokens that are propagated based on the thread's token sequence.}
  \vspace{-5mm}
\end{figure*}
\\
\indent
\NameSty{Thread} improves the flexibility of model generation, allowing the model to adapt, through recursive spawning, the amount of work it does based on the given problem, without overextending its context.
Importantly, this unfolds with limited need for explicit rules or hard coded logic. 
Instead, \NameSty{Thread} relies on the model's ability to infer the appropriate continuation given the context of each thread at the time of execution. 
Finally, \NameSty{Thread} is agnostic to the type of sequence. Depending on the underlying model, \NameSty{Thread} can be applied in varied settings, including multi-modal applications, 
and fulfill varied purposes (e.g., carrying out calculations, generating thoughts, retrieving information, robot manipulation). 
\\
\indent
In this paper, we consider \NameSty{Thread} in the settings of agent tasks (Figure 2a) and question answering (Figure 2b). In these settings, \NameSty{Thread} enables the LLM to recursively decompose a task, or question, into progressively simpler sub-problems that can be solved by child threads in a compartmentalized manner. A child thread infers the specifications of its sub-problem from its parent's token sequence. If needed, child threads can troubleshoot their sub-problem, and spawn their own threads, all without distracting their parent.  We test \NameSty{Thread}, using a few-shot learning approach, on benchmarks consisting of question-answering and agent tasks.
For these problems, we apply \NameSty{Thread} with join synchronization, allowing parent threads to see feedback from child threads before defining their next step.
For example, in the agent setting, if the model is given the task of cleaning a bowl and putting it on the countertop, the main thread may spawn a child with the sub-task of finding a bowl. This child can then spawn further threads for handling the complexities of navigating to and checking different locations for the bowl, with each thread returning only the information needed for its parent to proceed with its sub-task.
Based on the feedback with regard to finding the bowl, the main thread can spawn a new thread to re-attempt the sub-task or spawn threads to execute the remaining sub-tasks (i.e., washing the bowl, putting the bowl on the countertop).
\\
\indent
\NameSty{Thread} has several advantages over existing frameworks in these settings. To start, \NameSty{Thread} addresses limitations of methods that require the model to solve the problem in one line of context \cite{yao2023react, shinn2024reflexion, sun2023adaplanner} by allowing the model to dynamically offload work by spawning child threads.
Further, unlike methods for task decomposition \cite{khot2023decomposed, sun2023pearl,prasad2023adapt, wang2024tdag, wang2023plan}, \NameSty{Thread} enables the model to adapt its decision-making in real time as it receives feedback from each step, as depicted in Figure 2. Existing methods either do not allow for plan adaptation \cite{sun2023pearl, wang2023plan} or only adapt plans by calling pre-existing sub-task handlers \cite{khot2023decomposed} or by adding more sub-steps within the existing plan when a sub-task fails \cite{prasad2023adapt, wang2024tdag}.
Finally, \NameSty{Thread} provides a more unified framework compared to these methods, which require separate prompting mechanisms for separate planner and executor modules. Instead, \NameSty{Thread} 
can be implemented with the same few-shot prompt used for every thread at every step of the task completion.
\\
\indent
We evaluate \NameSty{Thread} on agent tasks and data-grounded question answering.
\NameSty{Thread} significantly outperforms prior methods, achieving state-of-the-art performance with GPT-4 and GPT-3.5 on ALFWorld, TextCraft, and WebShop, along with two new benchmarks, DataCommons QA and MIMIC-III ICU QA. \NameSty{Thread} also shows success with smaller models, outperforming prior methods by 10\% to 50\% absolute points across the benchmarks with Llama-3-8b and CodeLlama-7b.

\section{\NameSty{Thread} Framework}

\NameSty{Thread} frames model generation as an execution thread that, given the context, can run to completion or spawn new threads. A thread consists of generating with a model, $G$, given a specific starting context, $c$. For the main thread, $c$ is the initial seed context, $c_{0}$ (e.g., context provided by a user). For each child thread, $c$ derives from the parent's token sequence. A thread continues until it meets some termination criteria, such as a stop token.

\begin{algorithm}[H]
\caption{\NameSty{Thread} with join synchronization}
\begin{algorithmic}
\Function{Thread}{$c$, $Y$}
    \While{\texttt{True}}
        \State $Y=Y+G(c+Y)$
        \If{$Y$\text{ spawns a child thread}}
            \State $Y=Y+\, \psi(\,$\Call{Thread}{$\phi(Y), [\,]$\,}$\,)$
        \ElsIf{$Y$\text{ ends the thread}}
            \State \Return $Y$
        \EndIf
    \EndWhile
\EndFunction
\end{algorithmic}
\end{algorithm}

\subsection{\NameSty{Thread} with Join Synchronization}
We present \NameSty{Thread} in a setting where, analogous to how the \texttt{join()} method is used in multithreaded programming, a parent thread waits for the execution of the child to complete before proceeding. When the child execution completes, the child's output is returned directly to the parent. In this setting, \NameSty{Thread} can be implemented using the recursive function shown in Algorithm 1. The function takes two inputs: the context for the thread, $c$, and the tokens generated so far by the thread, $Y$. We treat a token sequence as a list of tokens. 

\NameSty{Thread} begins with $c = c_{0}$ and $Y=[\,]$. The model generates, $G(c+Y)$, one token at a time conditioning on the given context, $c$, appended with the growing sequence of generated tokens, $Y$.

If $Y$ spawns a child thread, then a new thread is created with \Call{Thread}{$\phi(Y), [\,]$\,} where $Y$ for the child thread is initialized as an empty list and $c$ is based on the token sequence of the parent, $\phi(Y)$. The output tokens of the child thread are appended to the parent's token sequence and the parent continues generating. 
If $Y$ ends the thread, then the thread returns $Y$ (its full token sequence). 

The $\phi$ function defines the context for a child thread based on the full token sequence of the parent thread at the time the child is spawned, including tokens directly generated by the parent or returned as output from previous child threads of that parent. The $\psi$ function defines the output tokens of a child thread based on its full token sequence at the time the thread ends, including tokens directly generated by the child or returned as output from threads that it spawned. 

\subsection{Alternative Synchronization and Spawning Mechanisms}
Above, we show \NameSty{Thread} in a setting where a parent always waits for a child to complete and the child's output is always appended directly to the token sequence of the parent before the parent proceeds. However, depending on the setting, \NameSty{Thread} can involve varied mechanisms for synchronization and spawning and, in scenarios without sequential dependencies, can include asynchronous multithreading, allowing parent threads to continue generating without waiting for child threads to complete (improving overall efficiency).

\subsection{Defining Parent-Child Information Exchange with \texorpdfstring{$\phi$}{Lg} and \texorpdfstring{$\psi$}{Lg}}
The functions $\phi$ and $\psi$ control the propagation of information from parent to child and from child to parent, respectively. Similar to the synchronization and spawning mechanisms mentioned above, these functions can vary based on the setting in which \NameSty{Thread} is implemented. For example, $\phi$ and $\psi$ can range from simple functions that compress or decompress information in the token sequence to entirely separate models that transform the sequence in more complex ways. 
In the section below, we describe how we define $\phi$ and $\psi$ when implementing \NameSty{Thread} in the settings of agent tasks and question answering.

\begin{figure*}
  \centering
  \includegraphics[width=0.8\textwidth]{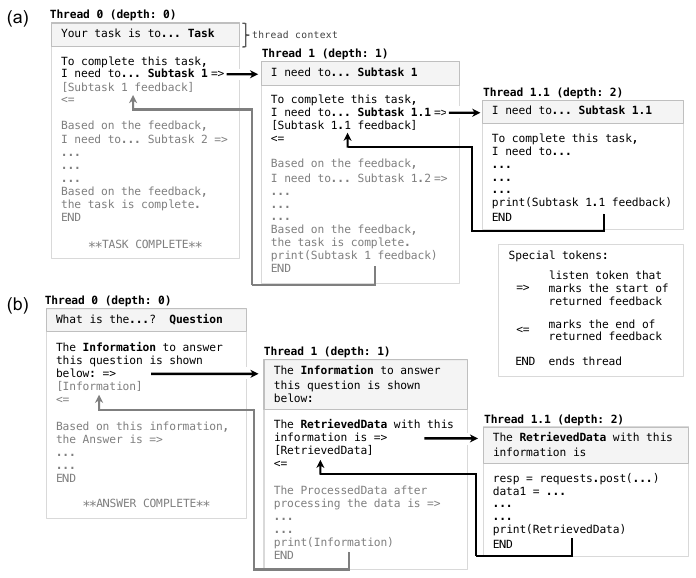}
  \caption{
  When given a task (a) or question (b), \NameSty{Thread} can be used to help the model, through recursive spawning, decompose the problem into progressively simpler sub-problems that are solved by child threads. In these examples, the context for a child thread is based on the last line of the parent's token sequence. \\ 
  \hfill 
  }
  \vspace{-8mm}
\end{figure*}

\subsection{Dynamically Adapting Intermediate Work in Token Generation} 
The \NameSty{Thread} framework enables a thread of model generation to spawn a child thread to produce output tokens that can serve as future tokens for the parent. In effect, this enables the model to adapt, as needed, the amount of work or computational steps used to produce different tokens. The intermediate work added through recursive spawning can range from deeper thinking and reasoning to interacting with external environments or sources of information. As depicted in Figure 3, the work associated with each part of the parent's token sequence is represented by a thread tree organized based on connections between threads. These trees reflect how threads interact to form output tokens and the computational work associated with each component of the overall model generation.

\section{Applying \NameSty{Thread} for Agent Tasks and Question Answering}
\label{section3}

In this paper, we test \NameSty{Thread} in the settings of question answering and agent tasks with LLMs, where \NameSty{Thread} enables the model to dynamically decompose the given problem into simpler sub-problems that are completed by separate threads. We apply \NameSty{Thread} using an in-context learning approach, where the same few-shot prompt is used for every thread at every step of the task completion or question answering. Since these problems benefit from the model adapting to feedback as it defines its next steps, we apply \NameSty{Thread} with the join synchronization described above.

\textbf{Thread spawning and termination based on special stop tokens.} We implement a spawning mechanism using a special stop token, $\omega_{listen}$, which pauses the thread generation and marks the start of the output the thread expects from the child. The context for the child thread is defined by $\phi$ based on the token sequence of the parent that occurs before $\omega_{listen}$. A thread ends when it generates the end token, $\omega_{end}$. As exemplified in Figure 2, we use \verb|=>| as $\omega_{listen}$ and \texttt{END} as $\omega_{end}$. 
\\
\\
\textbf{Implementing \NameSty{Thread} with few-shot learning.} We leverage a few-shot learning approach to implement \NameSty{Thread}. The few-shot examples are comprised in a single prompt with examples of successful spawning and problem solving at different thread depths, including how the token $\omega_{listen}$ should be used. As a result, the approach does not require explicit rules to define when new threads should be spawned and how parent threads should respond to feedback from a child. Instead, these mechanisms are all implied by the examples provided in the few-shot prompt. Changing the spawning and parent-child dynamics simply requires modifying the prompt. All threads are provided the same few-shot prompt for every step of the problem completion. Thus, \NameSty{Thread} only requires generating a single prompt for a given setting. This prompt, $q$, is prepended to the context for each thread, $G(q+c+Y)$, forming the full context from which the thread will generate.

\textbf{Defining $\phi$ and $\psi$}. As described above, the functions $\phi$ and $\psi$ control information flow from parent to child and from child to parent, respectively. The function $\psi$ returns the result from a child thread based on its full token sequence. As depicted in Figure 2, we implement $\psi$ as a function that returns the tokens included in the print statement at the end of the child thread's token sequence. In addition, $\psi$ appends the token \verb|<=| to the output to mark the end of the child's output within the parent's token sequence. We implement $\phi$ as a function that extracts the last line of the parent's token sequence to create the context for the child thread. To improve the ability of parent and child threads to efficiently organize, use, update, and propagate shared information, we leverage the coding-based skills of the model to define variables that represent pieces of information that are important for the given problem. For example, if the task is to find an item on an e-commerce site that has certain attributes, the model can define a variable, such as \verb|obj_attributes|, and spawn a child to search for items with \verb|obj_attributes| instead of having to list all of the attributes. Then, when $\phi$ processes the token sequence from the parent, it instantiates the variables, allowing the child to see, for example, the full list of attributes. We also test \NameSty{Thread} implemented without using these variables, with results shown in the appendix.

\textbf{Threads performing actions in an environment.} Threads can listen for feedback when performing an action in an environment the same way they listen for feedback from a child thread. They simply generate tokens that represent the action and then produce the token $\omega_{listen}$ to listen for feedback from the environment. The action, $a$, is then executed in the environment, $E$, and the output, $o = E(a)$, is appended to the thread's token sequence the same way it is done for the feedback from a child thread (Algorithm 2). The thread can then continue generating based on the feedback.

\section{Experiments}
We evaluate \NameSty{Thread} on 5 benchmarks for agent tasks and question answering using large and small models, including GPT-4, GPT-3.5, Llama-3-8b, Llama-2-7b, and CodeLlama-7b. For each benchmark, we use the same prompt for all models.

We include details regarding models, prompts, and experiments (including ablations) in the appendix. Examples of the prompts are shown in Appendix ~\ref{appendix_prompts}. Details regarding the QA benchmarks are provided in Appendix ~\ref{appendix_qa_benchmarks}. 
We release all code and data at \url{https://github.com/philipmit/thread}.

\subsection{ALFWorld}
ALFWorld is a suite of text-based environments, implemented in TextWorld, designed to align with the embodied ALFRED benchmark \cite{shridhar2021alfworld}. It includes 6 different task types that instruct the agent to accomplish a goal 
by interacting with a simulated household with actions defined in text. Following \citet{yao2023react}, we evaluate an agent on 134 unseen evaluation games, including six task types: Pick, Clean, Heat, Cool, Look, and Pick2. Like previous methods, such as ReAct \cite{yao2023react}, we implement \NameSty{Thread} with one few-shot prompt per task. We also test \NameSty{Thread} using task-general prompting.

\paragraph{Results.} \NameSty{Thread} significantly outperforms all prior methods (Tables 1 and 2), including those that require the agent to have access to external memory consisting of past experiences with the task. With GPT-4, \NameSty{Thread} achieves a combined success rate of 98.5\% with 100\% success in 4 of the 6 tasks. With GPT-3.5, \NameSty{Thread} shows a combined success rate of 95.5\%, outperforming all methods by over 9\% absolute points. With smaller models (Llama-3-8b and CodeLlama-7b), \NameSty{Thread} improves upon prior methods by 30\% to 55\% points.

Finally, Table 8 shows the results when testing \NameSty{Thread} with task-general prompting. To implement the task-general prompt, we split the prompt into one set of examples for the main thread and a second set of examples for all other threads (with the same sets of examples used for all tasks). We see that GPT-3.5 achieves the same combined success rate with the task-general prompting as it does with the task-specific prompting. Further, while the performance of Llama-3-8b and CodeLlama-7b degrades with task-general prompting, it remains a significant improvement over task-specific prompting with prior methods (Table 2).

\subsection{TextCraft}
TextCraft is a text-based environment inspired by the crafting component of Minecraft \cite{prasad2023adapt}. The tasks involve building Minecraft items with crafting commands using available resources from the environment. Following the work of \citet{prasad2023adapt}, we evaluate \NameSty{Thread} on the test set containing 200 samples. We implement \NameSty{Thread} using a single few-shot prompt and compare its performance to previous methods used for TextCraft \cite{prasad2023adapt, wang2024tdag, shinn2024reflexion, yao2023react}.

\begin{table*}
\caption{ \textbf{ALFWorld task-specific success rates (\%).} Results are separated based on whether the method requires the agent to access to external memory. *As reported in \citet{kagaya2024rap} (RAP) and \citet{fu2024autoguide} (AutoGuide). }
\centering
\resizebox{\textwidth}{!}{
\begin{tabular}{llllllllll}
\toprule
\textbf{Model} & \makecell[l]{\textbf{Requires} \\ \textbf{ext. mem.}} & \textbf{Method} & \textbf{All} & \textbf{Pick}	&\textbf{Clean} &	\textbf{Heat} &	\textbf{Cool}	&\textbf{Look}&	\textbf{Pick2} \\ 
\midrule

\multirow{9}{*}{GPT-3.5} 
& \multirow{5}{*}{Yes} & Reflexion \cite{shinn2024reflexion} & 76.1 & 75.0 & 80.6 & 69.6 & 76.2 & 83.3 & 70.6 \\ 
& & AdaPlanner \cite{sun2023adaplanner} & 82.8 & 91.7 & 87.1 & 82.6 & \textbf{95.2} & 50.0 & 82.4 \\ 
 & & RAP* \cite{kagaya2024rap} & 85.8 & 95.8 & 87.1 & 78.3 & 90.5 & 88.9 & 70.6  \\ 
 & & AutoGuide* \cite{fu2024autoguide} & 79.1 & - & - & - & - & - & - \\ 

\cmidrule(r){2-10}

 & \multirow{3}{*}{No} & ReAct \cite{yao2023react} & 53.7 & 45.8  & 48.4 & 69.6 & 66.7 & 55.6 & 35.3 \\ 
 & & DecomP \cite{khot2023decomposed}  & 84.3  & 91.7 & 87.1 & 82.6 & 90.5 & 83.3 & 64.7    \\ 

 & & ADaPT \cite{prasad2023adapt} & 82.1 & 87.5 & 83.9 & 78.3 & 90.5 & 83.3 & 64.7  \\ 

 & & THREAD & \textbf{95.5} & \textbf{95.8} & \textbf{93.5} & \textbf{95.7} & \textbf{95.2} & \textbf{100} & \textbf{94.1}  \\ 

\midrule

\multirow{5}{*}{GPT-4} 
& \multirow{1}{*}{Yes}  & RAP* \cite{kagaya2024rap} & 94.8 & 95.8  & 90.3 & \textbf{100} & \textbf{95.2} & \textbf{100} & 88.2 \\ 
\cmidrule(r){2-10}

 & \multirow{3}{*}{No} & ReAct \cite{yao2023react} & 87.3 & 83.3 & 77.4 & 95.7 & 85.7 & \textbf{100} & 88.2 \\ 
 & & DecomP \cite{khot2023decomposed}  & 89.6 & 87.5  & 87.1 & 91.3 & 90.5 & 94.4 & 88.2   \\ 
 & & ADaPT \cite{prasad2023adapt} & 91.0 & 91.7  & 87.1 & 95.7 & 90.5 & 94.4 & 88.2 \\ 
 & & THREAD & \textbf{98.5} & \textbf{100} & \textbf{100} & \textbf{100} & \textbf{95.2} & \textbf{100} & \textbf{94.1}  \\ 

\bottomrule
\end{tabular}
}
\end{table*}

\begin{wraptable}{l}{1\textwidth} 
\vspace{-5mm}
\caption{ \text{ALFWorld success rates (\%) for all tasks combined.}
}
\centering

\begin{tabular}{lllll}
\toprule
\makecell[l]{\textbf{Requires} \\ \textbf{ext. mem.}} & \textbf{Method} & \textbf{Llama-3-8b}  & \textbf{Llama-2-7b}	&\textbf{CodeLlama-7b}\\ 
\midrule

\multirow{3}{*}{Yes} & Reflexion \cite{shinn2024reflexion} & 25.4 & 11.2 &  27.6\\ 
& AdaPlanner \cite{sun2023adaplanner} & 28.4 & 11.9 & 29.9   \\ 

\midrule
 
\multirow{3}{*}{No} & ReAct \cite{yao2023react} & 20.1 & 12.7 & 23.1    \\ 

& DecomP \cite{khot2023decomposed}  & 37.3  & 14.2  &  33.6   \\ 
& ADaPT \cite{prasad2023adapt} & 30.6 & 15.7 & 35.1   \\ 

& THREAD  & \textbf{71.6} & \textbf{22.4} & \textbf{91.0}   \\ 

\bottomrule
\end{tabular}
\end{wraptable}
\hfill
\\
\hfill
\\
\hfill
\\
\hfill
\\
\hfill
\\
\hfill
\\
\hfill
\\
\hfill
\\
\hfill
\\
\hfill
\\
\hfill
\\
\hfill
\vspace{-5mm}

\paragraph{Results.}
\NameSty{Thread} outperforms prior methods by at least 20\% absolute points with GPT-3.5 and at least 40\% points with Llama-3-8b and 30\% points CodeLlama-7b (Table 3). Further, Llama-3-8b with \NameSty{Thread} outperforms GPT-3.5 with all prior methods and CodeLlama-7b with \NameSty{Thread} outperforms GPT-3.5 with all prior methods except TDAG.
\subsection{WebShop}
WebShop is an online shopping environment with over a million real-world products \cite{yao2022webshop}. The benchmark requires the model to interact with the website to purchase a product based on specifications provided by a user.
The evaluation metrics include the success rate, which is the percentage of products that were purchased with full success, and the score, which is the average percentage of desired attributes covered by the purchased items. Following prior work \cite{shinn2024reflexion, prasad2023adapt, zhou2023language}, we evaluate \NameSty{Thread} on a test set of 100 instructions. 
\hfill
\\
\hfill
\\
\hfill
\\
\hfill
\\
\hfill
\\
\hfill
\\
\hfill
\\
\hfill
\\
\hfill
\\
\hfill
\vspace{4mm}

\begin{table*}
\caption{TextCraft success rate (\%). *As reported in \citet{prasad2023adapt}.}
\centering
\begin{tabular}{lllll}
\toprule
\multirow{1}{*}{\textbf{Method}} & \multicolumn{1}{c}{\text{GPT-3.5}} & \multicolumn{1}{c}{\text{Llama-3-8b}} &	\multicolumn{1}{c}{\text{Llama-2-7b}} & \multicolumn{1}{c}{\text{CodeLlama-7b }}   \\ 
\midrule

Reflexion* \cite{shinn2024reflexion} & 32.0 & - & - & - \\ 

ReAct \cite{yao2023react} & 20.5 & 12.5 & 8.0 & 10.5  \\ 

DecomP \cite{khot2023decomposed}  & 68.5  &  47.0   & 12.0  & 38.5  \\

ADaPT  \cite{prasad2023adapt}& 52.5 & 23.5  & 12.0 & 18.0 \\ 

TDAG  \cite{wang2024tdag} & 73.5 & 48.5  & 14.0 & 31.0 \\ 

THREAD & \textbf{93.5} & \textbf{92.0} & \textbf{20.0} & \textbf{71.0} \\ 

\bottomrule
\end{tabular}
\end{table*}

\begin{table*}
\caption{WebShop success rate (SR; \%) and score (\%).}
\centering
\resizebox{\textwidth}{!}{
\begin{tabular}{llllllllll}
\toprule
\multirow{2}{*}{\makecell[l]{\textbf{Requires} \\ \textbf{ext. mem.}}} & \multirow{2}{*}{\textbf{Method}} & \multicolumn{2}{c}{\text{GPT-3.5}} & \multicolumn{2}{c}{\text{Llama-3-8b}}  &	\multicolumn{2}{c}{\text{Llama-2-7b}}&	\multicolumn{2}{c}{\text{CodeLlama-7b }} \\ 
\addlinespace[2pt]
 & & \textbf{SR} & \textbf{Score} & \textbf{SR \; \; } & \textbf{Score} & \textbf{SR} & \textbf{Score} & \textbf{SR} & \textbf{Score} \\
\midrule

\multirow{5}{*}{Yes} & Reflexion \cite{shinn2024reflexion} & 38 & 64.4 & 32 & 59.8 & 8 & 19.7 & 17 & 57.3 \\ 
 & LATS \cite{zhou2023language} & 40 & 76.0 & 34 & 61.5 & 12 & 30.1 & 21 & 60.7 \\ 
 & RAP* \cite{kagaya2024rap} & 48 & 76.1  & - & - & - & - & - & -  \\ 
 & AutoGuide* \cite{fu2024autoguide} & 46 & 73.4 & - & - & - & - & - & -  \\ 

\cmidrule(r){1-10}

 \multirow{4}{*}{No} & ReAct \cite{yao2023react} & 37 & 59.5  & 31 & 54.1 & 10 & 18.5 & 17 & 49.2 \\ 

                     & DecomP \cite{khot2023decomposed}  & 43  &  58.7 & 35  &  56.3   & 14  & 29.6  & 21  & 57.1 \\

                     & ADaPT \cite{prasad2023adapt}  & 44 & 60.0 & 35 &  58.2  & 13 & 28.7 & 21 & 56.4 \\ 

                      & TDAG \cite{wang2024tdag} & 45 & 64.5 & 37 &  63.5  & 14  & 29.8 & 23 & 58.5 \\ 

 & THREAD & \textbf{49} & \textbf{76.3} & \textbf{47} & \textbf{70.4} & \textbf{20} & \textbf{48.5}  & \textbf{40} & \textbf{68.9} \\ 

\bottomrule
\end{tabular}
}
\end{table*}

\hfill
\paragraph{Results.}
Table 4 shows the success rate and score for each method. We again separate the results based on whether the method requires the model to have access to external memory. With GPT-3.5, \NameSty{Thread} outperforms other prompt-only methods by an absolute 4\% in success rate and over 10\% in score and outperforms RAP \cite{kagaya2024rap} by 1\% in success rate (with a similar score). \NameSty{Thread} achieves an absolute 10\%, or greater, improvement in success rate with Llama-3-8b and CodeLlama-7b. Llama-3-8b with \NameSty{Thread} achieves a higher success rate than GPT-3.5 with all prior methods, with the exception of RAP.

\subsection{DataCommons QA}

DataCommons QA is a benchmark consisting of questions that can be answered using data provided by Google DataCommons \cite{guha2019datacommons} \url{https://datacommons.org}. The questions range from comparing statistics in  different locations to making predictions regarding future trends. 
The test set includes a total of 140 questions. We implement \NameSty{Thread} with a few-shot prompt and compare its performance to three baselines: Reflexion \cite{shinn2024reflexion}, Natural Language Embedded Programs (NLEP) \cite{zhang2023natural}, and NLEP+ReAct, an extension of NLEP that allows the model to evaluate intermediate outputs of its analysis as it answers the question.

\paragraph{Results.}
Table 5 shows the accuracy of each method on DataCommons QA. \NameSty{Thread} outperforms prior methods by over 10\% absolute points with GPT-3.5, Llama-3-8b, and CodeLlama-7b. Llama-3-8b with \NameSty{Thread} outperforms GPT-3.5 with all prior methods.

\begin{table*}
\caption{DataCommons QA accuracy (\%).}
\centering
\begin{tabular}{lllll}
\toprule
\multirow{1}{*}{\textbf{Method}} & \multicolumn{1}{c}{\text{GPT-3.5}} & \multicolumn{1}{c}{\text{Llama-3-8b }} &	\multicolumn{1}{c}{\text{Llama-2-7b}} & \multicolumn{1}{c}{\text{CodeLlama-7b }}\\ 

\midrule

Reflexion \cite{shinn2024reflexion}      & 37.9  & 24.3          & 10.7          & 20.7  \\ 
DecomP \cite{khot2023decomposed}         & 64.3  & 57.9          &        15.7    &  31.4  \\ 

NLEP \cite{zhang2023natural}          & 28.6  & 21.4          & 11.4          & 19.3  \\ 
NLEP+ReAct     & 41.4  & 27.1          & 14.3          & 24.3\\ 
THREAD & \textbf{77.1} & \textbf{67.9} & \textbf{22.1} & \textbf{62.1} \\ 
\bottomrule
\end{tabular}
\end{table*}

\begin{table*}
\caption{MIMIC-III ICU QA accuracy (\%).}
\centering
\begin{tabular}{lllll}
\toprule
\multirow{1}{*}{\textbf{Method}} & \multicolumn{1}{c}{\text{GPT-3.5}} & \multicolumn{1}{c}{\text{Llama-3-8b }}  &	\multicolumn{1}{c}{\text{Llama-2-7b}}  & \multicolumn{1}{c}{\text{CodeLlama-7b }}\\ 

\midrule

Reflexion \cite{shinn2024reflexion} & 38.1  & 19.4 &  8.8 & 16.9   \\ 
DecomP \cite{khot2023decomposed}          & 61.9  & 51.3          &    13.1        &  30.6  \\ 

NLEP \cite{zhang2023natural}  & 35.6 & 17.5 & 8.1 & 15.6\\ 
NLEP+ReAct & 43.1 & 23.8 & 12.5  & 21.9\\ 
THREAD & \textbf{71.3} & \textbf{61.9} & \textbf{18.8}  & \textbf{58.1} \\ 
\bottomrule
\end{tabular}
\end{table*}

\subsection{MIMIC-III ICU QA}
The MIMIC-III ICU QA benchmark consists of patient-focused questions based on clinical time-series data made available by MIMIC-III \cite{johnson2016mimic}. The benchmark reflects a setting in which a healthcare provider can ask natural language questions about patients in the intensive care unit (ICU) 
and receive an answer from the language model based on the relevant patient data. 
The test set includes 160 questions. Similar to above, we implement \NameSty{Thread} with a few-shot prompt and compare its performance relative to Reflexion \cite{shinn2024reflexion}, NLEP, and NLEP+ReAct.

\paragraph{Results.}
As shown in Table 6, \NameSty{Thread} significantly outperforms prior methods across all models except Llama-2-7b. All methods show the lowest performance with Llama-2-7b, which is consistent across all benchmarks.

\section{Related Work}

The \NameSty{Thread} framework relates to prior approaches for dynamically adapting work, or computational steps, used during model generation. Previous approaches for dynamically adapting computation based on the given problem require special training \cite{goyal2023think, nye2022show} or novel model architectures \cite{graves2016adaptive, banino2021pondernet, dehghani2018universal}. \NameSty{Thread} provides a more general and flexible framework that, as we show, can be applied without modifications to the underlying model architecture and without further training.  In addition, unlike methods that allow the model to adapt the amount of internal computations \cite{goyal2023think, graves2016adaptive, banino2021pondernet, dehghani2018universal}, the intermediate work used to supplement model generation with \NameSty{Thread} can involve work that extends beyond internal thinking, such as retrieving information or interacting with an external environment. In addition, \NameSty{Thread} allows for greater interpretability, since all of the intermediate work is performed by generating meaningful tokens.
\\
\indent
In this paper, we apply \NameSty{Thread} in a setting where it improves the ability of LLMs to decompose a given problem into simpler sub-problems. There has been significant prior work involving the decomposition of problems with neural modular modeling architectures \cite{andreas2016neural, talmor2018web, min2019multi, jiang2019self, gupta2019neural, perez2020unsupervised, khot2021text}. Later work has used fine-tuning or in-context learning with modern LLMs for decomposition with multi-step tasks \cite{khot2023decomposed, wang2023plan}, mathematical reasoning \cite{gao2023pal, lee2023recursion}, and program synthesis \cite{murali2018neural, nye2019learning, zheng2023outline}. To handle complex tasks, a more recent approach, called ADaPT, has improved upon these methods by using a planner to further decompose a task when a failure is encountered \cite{prasad2023adapt}, which was further extended to a multi-agent setting with TDAG \cite{wang2024tdag}. Unlike prior methods, such as Decomposed Prompting \cite{khot2023decomposed}, ADaPT and TDAG enable the model to adapt a plan upon task failure. In contrast to these approaches, \NameSty{Thread} has the advantage of allowing the model to adapt its decision-making in real time as it receives feedback from each step. With \NameSty{Thread}, each step of the plan is specified only after receiving feedback from the previous step. In addition, \NameSty{Thread} enables the model to flexibly specify and, if needed, re-specify sub-tasks when a sub-task fails or when unexpected feedback is returned. Finally, \NameSty{Thread}, which uses the same prompting for every thread at every step of the problem, provides a more unified framework than ADaPT and TDAG, which involve separate prompting and few-shot examples for the “planner” and “executor” modules. With these methods, the planner module outlines a full plan and the executor follows all steps of the plan. If the executor fails, the planner is tasked with further decomposing the failed step into more sub-steps. Instead of outlining a full plan and attempting all steps, \NameSty{Thread} involves specifying one step, allocating this to a child thread, and receiving feedback from the child thread before proceeding. 
\section{Conclusion}
We introduce \NameSty{Thread}, a general framework in which model generation is treated as a thread of execution that can dynamically offload work, such as thinking or retrieving information, by spawning new threads. \NameSty{Thread} enables a model to adapt, through recursive spawning, the amount of intermediate work it uses to produce different parts of its token sequence. We apply this framework in the settings of agent tasks and question answering. We show that \NameSty{Thread}, implemented using a few-shot learning approach, achieves state-of-the-art performance on diverse benchmarks.

\section*{Acknowledgements}
This work was supported by the MIT-IBM Watson AI Lab and Quanta Computer, Inc. We also thank the Google DataCommons team for their help with developing the DataCommons QA benchmark.

\section*{Limitations}
The implementation of \NameSty{Thread} in this paper does not involve any explicit mechanisms for error handling and instead relies on the inherent reasoning ability of LLMs to respond to unexpected feedback from child threads or from the environment. An advantage of \NameSty{Thread} is that individual threads can fail without directly affecting other threads. However, the context necessary to self-correct may be lost if error reporting is not handled appropriately. More work is needed to develop robust error detection and recovery mechanisms to ensure that the information necessary for self-correction is preserved and utilized effectively. In addition, our implementation of \NameSty{Thread} involves limited communication between the parent and child threads, where the context for the child thread is based on the last line of the parent's token sequence. This can result in the child missing information that could help with its work, leading to less efficient or effective generation. Overall, more work is need to improve the propagation of information, as defined by the functions $\phi$ and $\psi$, between parent and child threads.

\section*{Ethical Statement}
In this work, we show how \NameSty{Thread} can be applied to improve LLM agent task completion and question answering. LLMs are being increasingly deployed to autonomously interact with external environments and humans. By improving this capacity, our work has the potential for amplifying risk associated with automated decision-making or facilitate harmful use of LLMs. Addressing these risks requires careful consideration of ethical guidelines, robustness checks, and transparency in algorithm deployment to ensure that advancements in LLMs contribute positively to societal welfare.



\newpage
\hfill

\begin{figure*}
  \centering
  \includegraphics[width=0.95\textwidth]{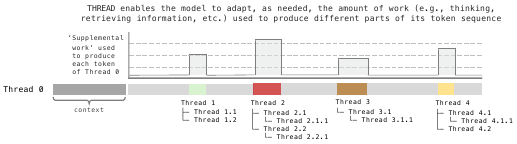} 
  \caption{\NameSty{Thread} allows model to adapt amount of supplemental work used to produce tokens.}
\end{figure*}

\newpage

\appendix

\section{Illustration of Intermediate Work used to Produce Tokens} 
Figure 3 illustrates how \NameSty{Thread} enables the model to dynamically adapt the amount of work or computational steps used to produce different parts of its token sequence. The illustration is based on the Thread 0 from Figure 1. The ``supplemental work'' for a token reflects the amount of additional tokens, generated by spawned threads, used to produce each token within Thread 0.  The work associated with each part of the token sequence is represented by a thread tree organized based on connections between parent and child threads. These trees reflect how threads interact to form output tokens and the computational work associated with each component of the overall model generation.

\section{Implementing \NameSty{Thread} for Agent Tasks and Question Answering}
We release all code and data at \url{https://github.com/philipmit/thread}. As described in section ~\ref{section3}, we implement \NameSty{Thread} with join synchronization for the problems evaluated in this paper. We implement a spawning mechanism using a special stop token, $\omega_{listen}$. A thread ends when it generates the end token, $\omega_{end}$. We use \verb|=>| as $\omega_{listen}$ and \texttt{END} as $\omega_{end}$. 

We implement \NameSty{Thread} in each setting using a fixed few-shot prompt showing examples of successful spawning and problem solving at different thread depths. All threads are provided the same few-shot prompt for every step of the problem completion. This prompt, $q$, is prepended to the context, $c=q+c$, for each thread, forming the full context from which the thread generates. 

As shown in Algorithm 2 below, threads can get feedback from the environment the same way they get feedback from a child thread. They generate tokens that represent the action and then produce the token $\omega_{listen}$ to listen for feedback from the environment. As shown in Algorithm 2, the action, $a$, is executed in the environment, $E$, and the output, $o = E(a)$, is appended to the thread's token sequence before it continues generation. The function $\xi$ parses the action from the token sequence. As with prior methods such as ReAct \cite{yao2023react}, actions are specified following the `>' token.

\noindent
\begin{algorithm}[H]
\caption{\NameSty{Thread} with join synchronization for
tasks involving actions in environment}
\begin{algorithmic}
\Function{Thread}{$c$, $Y$}
    \While{\texttt{True}}
        \State $Y=Y+G(c+Y)$
        \If{$Y$\text{ performs an action}}
            \State $a = \xi(Y)$
            \State $o = E(a)$
            \State $Y=Y+o$
        \ElsIf{$Y$\text{ spawns a child thread}}
           \State $Y=Y+\, \psi(\,$\Call{Thread}{$\phi(Y), [\,]$\,}$\,)$
        \ElsIf{$Y$\text{ ends the thread}}
            \State \Return $Y$
        \EndIf
    \EndWhile
\EndFunction
\end{algorithmic}
\end{algorithm}

To ensure consistency across benchmarks and with prior work, we use \texttt{Meta-Llama-3-8B} for Llama-3-8b, \texttt{Llama-2-7b-hf} for Llama-2-7b, and \texttt{CodeLlama-7b-hf} for CodeLlama-7b available on Huggingface \cite{wolf2019huggingface} and use \texttt{gpt-4-0613} for GPT-4 and \texttt{gpt-3.5-turbo-instruct} for GPT-3.5 from the OpenAI API, with the temperature set to 0 for all experiments.

\begin{table*}
\caption{Mean and standard error of \NameSty{Thread} performance and thread depths with GPT-3.5. }
\centering
\begin{tabular}{llll}
\toprule
\multirow{1}{*}{\textbf{Benchmark}} & Mean (Std. Err.) & Avg. thread depth & Max thread depth\\ 
\midrule
DataCommons QA & 76.7 (.17) & 4.1 & 6  \\ 

MIMIC-III ICU QA & 71.4 (.36) & 4.3 & 6 \\ 

ALFWorld & 95.7 (.28) & 3.7 & 7 \\ 

TextCraft & 93.7 (.37) & 5.8 & 10  \\ 

WebShop & 48.6 (.40) & 3.6 & 7 \\ 
\bottomrule
\end{tabular}
\end{table*}



\begin{table*}
\caption{ \text{ALFWorld task-specific success rates (\%) using task-general prompting with \NameSty{Thread}.} }
\centering
\begin{tabular}{llllllll}
\toprule
\textbf{Model} & \textbf{All} & \textbf{Pick}	&\textbf{Clean} &	\textbf{Heat} &	\textbf{Cool}	&\textbf{Look}&	\textbf{Pick2} \\ 
\midrule

\multirow{1}{*}{GPT-3.5} 
   & 95.5 & 100 & 96.8 & 82.6 & 95.2 & 100 & 100 \\  
\multirow{1}{*}{Llama-3-8b} 
   & 49.3 & 58.3 & 38.7 & 56.5 & 57.1 & 72.2 & 11.8 \\ 
\multirow{1}{*}{CodeLlama-7b} 
   & 61.9 & 41.7 & 87.1 & 65.2 & 61.9 & 38.9 & 64.7 \\ 
\bottomrule
\end{tabular}
\end{table*}

\section{Additional Experiments with GPT-3.5}
Table 7 shows the mean and standard error of the performance of \NameSty{Thread} with GPT-3.5 across 5 runs on each benchmark. Overall, the results are consistent across multiple runs. Table 7 also shows the average and max thread depths for each task.

\section{Task-general Prompting for ALFWorld}

Table 8 shows the results when testing \NameSty{Thread} with task-general prompting in ALFWorld. To implement the task-general prompt, we split the prompt into one set of examples for the main thread and a second set of examples for all other threads. The same sets of examples used for all tasks. GPT-3.5 achieves the same combined success rate with the task-general prompting as it does with the task-specific prompting. Further, while the performance of Llama-3-8b and CodeLlama-7b degrades with task-general prompting, it remains a significant improvement over task-specific prompting with prior methods (Table 2).

\section{Error and Ablation Analysis}
\subsection{Error Types}
To identify what types of errors \NameSty{Thread} reduces, we classify the errors that are responsible for each method’s failures. Descriptions and examples of each error type are provided in Tables 9 and 10 for the agent tasks and Tables 11 and 12 for the QA tasks. We classified these errors through hand review of each failure case, where the first error type to occur within the failure case was identified as the error type responsible for the failure. Figures 4a, 4b, 5, and 6a show the error counts across different methods and tasks. We focus this analysis on the prompt-only methods to provide a more clear evaluation of how the novel aspects of \NameSty{Thread} change performance relative to other methods that leverage few-shot learning. 

\subsection{Ablations}
Figures 4c, 4d, and 6b show how the error counts change when applying the following three modifications to \NameSty{Thread}:
\begin{enumerate}
    \vspace{-2mm}
    \item Removing the variables used by parent and child threads to manipulate and organize shared information.
    \vspace{-2mm}
    \item Replacing $\phi$ with a function that returns the parent thread’s full token sequence.
    \vspace{-2mm}
    \item Using a separate prompt for Thread 0 (which performs the initial task decomposition) instead of using the same few-shot prompt for all threads.
\end{enumerate}

\subsection{Agent Tasks}
\textbf{Task decomposition and flexible sub-task specification reduce failures caused by inadequate plans.}
In Figures 4a and 4b, we see that the methods that involve task decomposition (i.e., Decomposed prompting, ADaPT, and \NameSty{Thread}) show fewer failures that are caused by inadequate planning. This is likely because there is a dedicated line of reasoning for generating the plan, instead of it being part of the single line of model generation that is responsible for both defining the plan and carrying out the steps of the plan, like in ReAct. 

We see that \NameSty{Thread}, which allows for flexible sub-task specification, further reduces failures caused by inadequate planning. This is due to the fact that, when a child thread fails, the parent thread can accommodate the child thread’s feedback, adapt the sub-task specification, and spawn a new thread. Prior methods do not allow for this flexible sub-task re-specification, as they either require the model to select from a fixed set of pre-defined sub-task handlers (e.g., Decomposed Prompting) or do not accommodate sub-task feedback when handling failures (e.g, ADaPT and TDAG). 
Figure 7 shows an example where a child thread, which is tasked with checking a cabinet for a plate, returns feedback saying the cabinet is closed. This is a common problem, where a child thread (or a sub-task handler/executor in Decomposed Prompting or ADaPT) return some unpredictable intermediate result instead of fully completing its task (which, in this case, involves opening the cabinet and checking for a plate). To address this with \NameSty{Thread}, the parent thread spawns a new child thread with a modified sub-task specification (which includes opening the cabinet and checking if there is a plate) in order to prevent the new child thread from repeating the mistake made by the previous child. Prior methods do not allow for these dynamic, nuanced adjustments to the plan to accommodate the full range of possible intermediate feedback involved in complex tasks.

\textbf{Shortening action sequences reduces failures caused by misinterpreting environment feedback.}
We see that errors in interpreting environment feedback are lowest with Decomposed Prompting and \NameSty{Thread}. One potential explanation for this is that, by decomposing the task into more sub-components than other methods, Decomposed Prompting and \NameSty{Thread} require each line of model generation to execute shorter sequences of consecutive actions. As a result, each line of model generation needs to manage shorter action-observation sequences and, therefore, is able to more accurately interpret and keep track of the information returned from the environment.

Figure 5 shows the number of times GPT-3.5 misinterprets environment feedback when executing action sequences of different lengths when using the different methods. By significantly reducing the number of cases where the model has to perform long action sequences, Decomposed Prompting and \NameSty{Thread} reduce the number of interpretation errors.

\subsection{QA Tasks}
\textbf{Task decomposition reduces failures that are caused by performing the wrong analysis.}
In Figure 6a, we see that Decomposed Prompting and \NameSty{Thread} show fewer failures caused by the model performing the wrong analysis for the given question. This is similar to what we see with the agent tasks where problem decomposition improves the planning. By separating the line of model generation that defines the plan (defining the analysis type) from the line of model generation that is responsible for executing the plan (completing the analysis), the model can more effectively accomplish both tasks.

\textbf{Flexible sub-task specification reduces failures that are caused by runtime errors.}
We see that \NameSty{Thread} shows the fewest failures that are caused by runtime errors. With \NameSty{Thread}, when an error occurs, the error is returned to the parent thread, which can adapt its plan by modifying the sub-task specification and spawning a new thread (as discussed above). In addition, based on the ablation analysis (Figure 6b), we see that the variables used in \NameSty{Thread} may also reduce failures due to runtime errors, especially with smaller models such as Llama-3-8b.

\begin{figure*}
  \centering
  \includegraphics[width=1\textwidth]{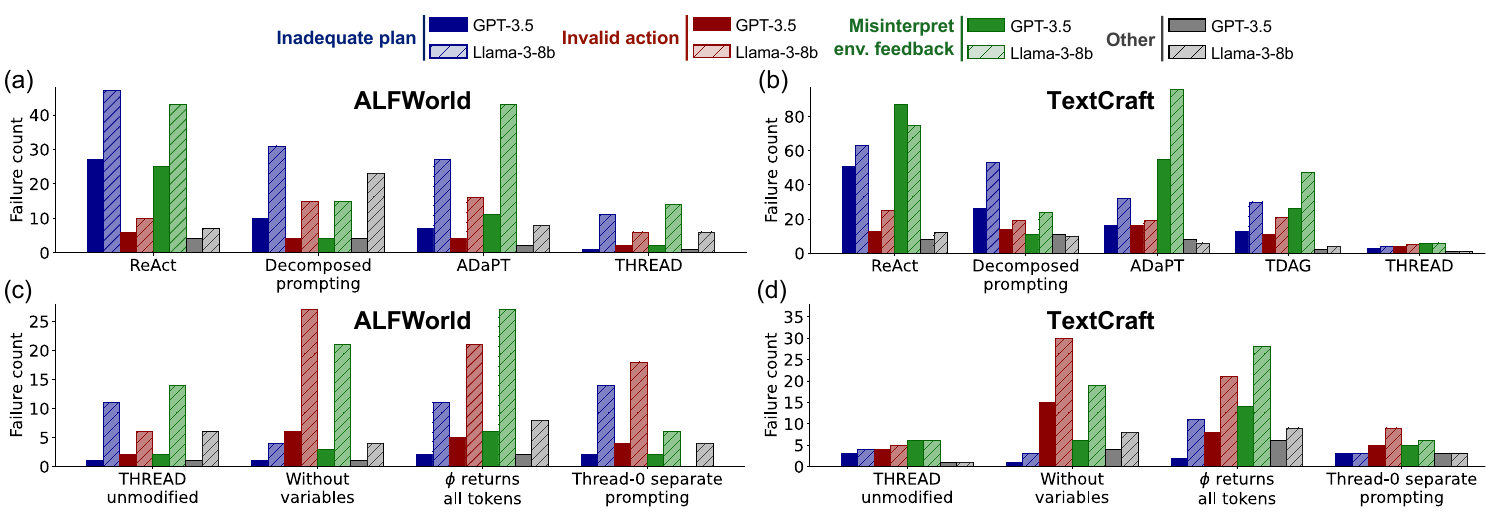}
  \caption{Failure counts in ALFWorld and TextCraft for different methods (a and b) and for modified versions of \NameSty{Thread} (c and d). }
  \vspace{-5mm}
\end{figure*}

\begin{figure*}
  \centering
\includegraphics[width=1\textwidth]{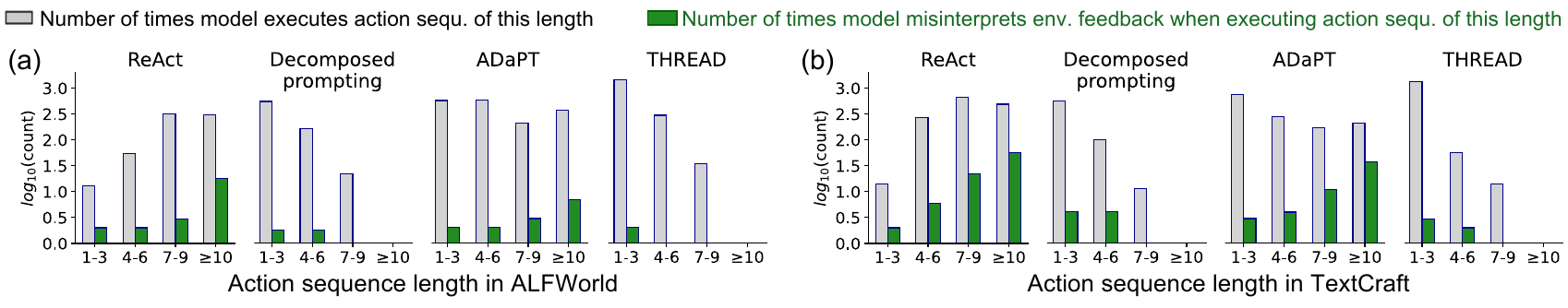}
  \caption{Number of times GPT-3.5 misinterprets environment feedback when executing action sequences in ALFWorld (a) and TextCraft (b). Values are shown as $\text{log}_{10}\text{(count)}$ to allow the total counts (shown in gray) to fit in the same figure as the error counts (shown in green). }
  \vspace{-5mm}
\end{figure*}

\begin{figure*}
  \centering
\includegraphics[width=1\textwidth]{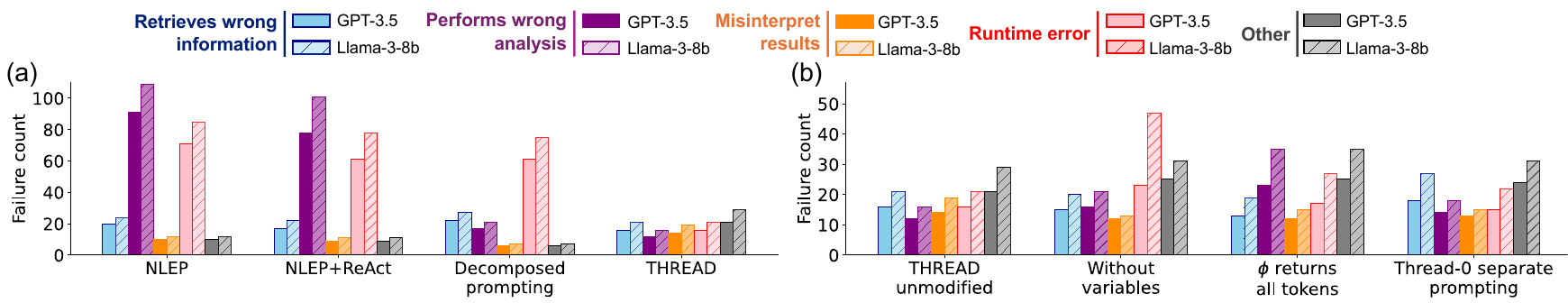}
  \caption{Failure counts, combined for DataCommons QA and MIMIC-III ICU QA, for different methods (a) and modified versions of \NameSty{Thread} (b). }
  \vspace{-5mm}
\end{figure*}

\begin{table*}
\caption{Description of error types for agent tasks.}
\centering
\begin{tabular}{p{4.5cm}p{10.5cm}} 
\toprule
\textbf{Error type} & \textbf{Description} \\
\midrule
\raggedright Inadequate plan & The plan does not include steps sufficient to complete the task (including steps to recover from unexpected intermediate results). \\ \\
\raggedright Invalid action &  The model attempts to perform an invalid action in the environment. \\ \\
\raggedright Misinterprets environment feedback & The model incorrectly interprets information returned from the environment. \\
\bottomrule
\end{tabular}
\end{table*}

\begin{table*}
\caption{Examples of error types for agent tasks.}
\centering
\begin{tabular}{p{4.5cm}p{10.5cm}} 
\toprule
\textbf{Error type} & \textbf{Examples} \\
\midrule
\raggedright Inadequate plan & \textbf{ALFWorld:} The plan does not include cleaning the plate when the model is tasked with putting a clean plate on the table. The plan does not involve opening the cabinet to check for a plate when an unexpected sub-task result is returned (e.g., “The cabinet is closed”). \textbf{TextCraft:} The plan involves crafting an object before acquiring one of the precursors. The plan does not finish checking if a material (e.g., bamboo) can be retrieved from the environment when an unexpected result is returned (e.g., “Could not find valid recipe for bamboo.”). 
\\ \\
\raggedright Invalid action & \textbf{ALFWorld:} The model incorrectly specifies the action for examining an object with the desklamp. \textbf{TextCraft:} The model incorrectly states a crafting recipe. 
\\ \\
\raggedright Misinterprets environment feedback & \textbf{ALFWorld:} The model continues looking for a plate after the environment indicates “You see a plate 1”. \textbf{TextCraft:} The model proceeds as though it has successfully fetched bamboo despite the environment returning “Could not find bamboo”. 
\\ \\
\raggedright Other & \textbf{ALFWorld:} The model reaches the maximum number of actions allowed in the ALFWorld environment. \textbf{TextCraft:} The model runs out of context. 
\\
\bottomrule
\end{tabular}
\end{table*}

\begin{table*}
\caption{Description of error types for QA tasks.}
\centering
\begin{tabular}{p{4.5cm}p{10.5cm}} 
\toprule
\textbf{Error type} & \textbf{Description} \\
\midrule
\raggedright Retrieves wrong information & The model does not retrieve the information needed to answer the question. \\ \\
\raggedright Performs wrong analysis & The model retrieves the correct information, but does not perform the analysis needed to answer the question. \\ \\
\raggedright Misinterprets results of analysis & The model performs the correct analysis, but misinterprets the results. \\ \\
Runtime error & Error occurs during the execution of the model's code. \\
\bottomrule
\end{tabular}
\end{table*}

\begin{table*}
\caption{Examples of error types for QA tasks.}
\centering
\begin{tabular}{p{4.5cm}p{10.5cm}} 
\toprule
\textbf{Error type} & \textbf{Examples} \\
\midrule
\raggedright Retrieves wrong information & The model retrieves data on smoking prevalence of all people when asked specifically about the female population. \\ \\
\raggedright Performs wrong analysis & The model compares values from 2014 to 2021 when asked to compare values from 2018 to 2021. \\ \\
\raggedright Misinterprets results of analysis & The model indicates a variable is increasing despite its analysis showing that it is decreasing. \\ \\
\raggedright Runtime error & The model tries to concatenate an integer to a string. \\ \\
\raggedright Other & The model runs out of context. \\
\bottomrule
\end{tabular}
\end{table*}

\begin{figure*}
  \hspace*{0pt}
\includegraphics[width=1\textwidth]{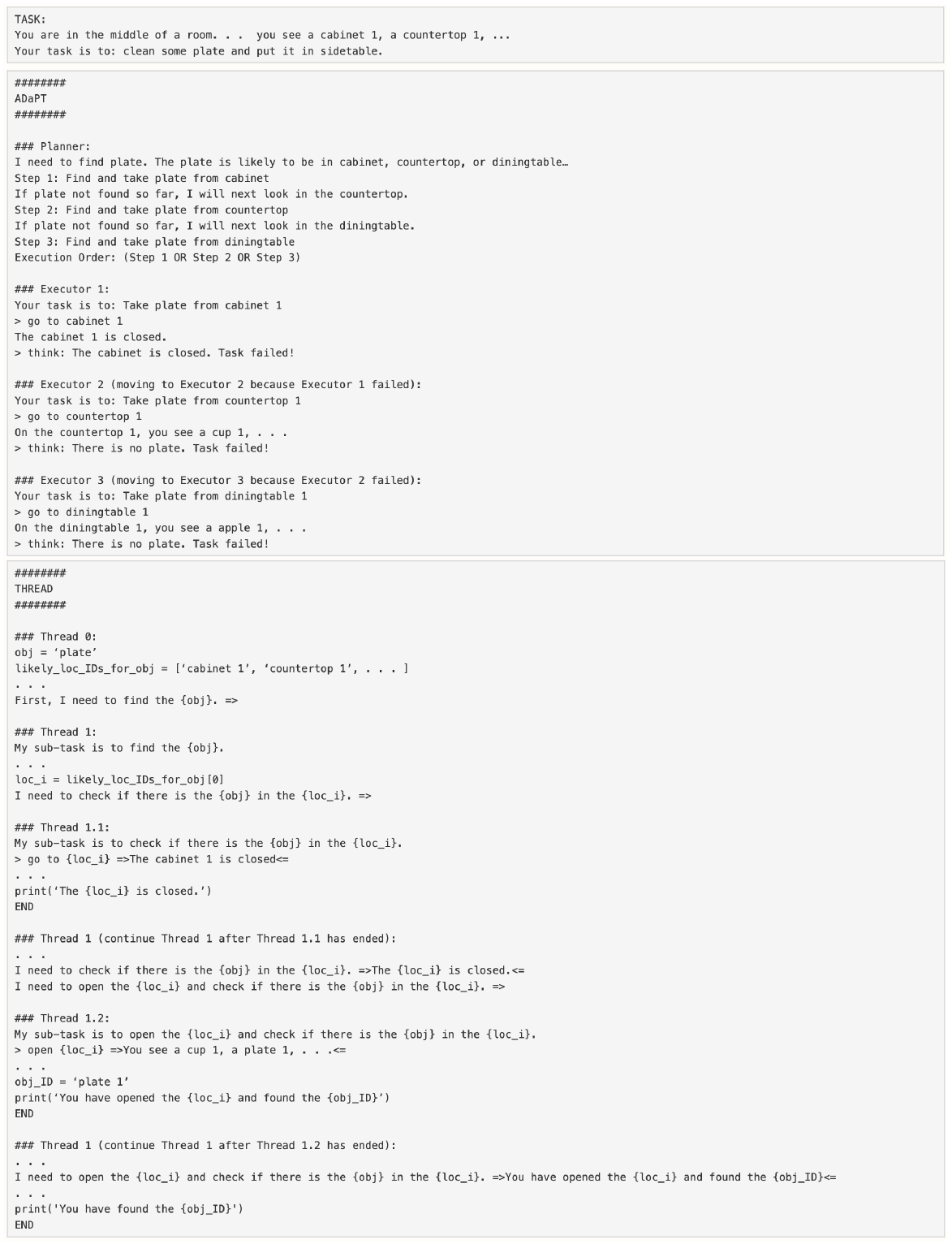}
  \caption{Example of task in ALFWorld where ADaPT fails to productively respond to the feedback from a sub-task, while  \NameSty{Thread} successfully accommodates the feedback to re-specify a new sub-task. }
\end{figure*}

\section{DataCommons QA and MIMIC-III ICU QA Benchmarks}
\label{appendix_qa_benchmarks}
We built the DataCommons QA benchmark utilizing data provided by Google Data Commons \cite{guha2019datacommons} with a focus on U.S. data. Data Commons contains publicly available data, aggregated from sources such as the Center for Disease Control and Prevention and the World Health Organization. 
\\
\indent
Generation of the benchmark started with the production of questions which fit the following criteria: 1) answerable by information provided by Google Data Commons, 2) broad enough to encompass information sampled between different locations (e.g., two counties), 3) lacking in subjectivity for the creation of a verifiable answer. The benchmark consists of question templates, which generate the questions of the benchmark, and ground-truth programs, which generate the answers to the questions of the benchmark. This format allows for the creation of a large quantity of diverse questions. Each question template consists of a sentence containing placeholders such as \verb|{variable_text}| which, upon random sampling of a variable, are altered accordingly, facilitating the automated production of a variety of questions based on one question template. An example of a  question template is as follows:
'Was \verb|{variable_text}| in \verb|{entity1}| increasing or decreasing from \verb|{time1}| to \verb|{time2}|?'. Time series data was sampled at the city, state, county, and country level for the United States. Data retrieval from Google Data Commons was conducted through the REST API provided by the website. 
\\
\indent
A ``ground-truth program'' was created for each question template to generate a verifiably correct answer for each question within the benchmark. These programs varied in length from 100-200 lines of code and analyzed data acquired from the Google Data Commons API in accordance with the constraints tailored by the question it was answering. For example, if a question template included a constraint about the years of data to be sampled, the ground-truth program was constructed to generate a year range for the answer. Code within the program would subsequently be created to filter and analyze the data acquired from Google Data Commons accordingly by applying whichever mathematical concept was required for the answer (e.g., correlation, linear regression, median, mean). Consequently, ground-truth programs formed the backbone of the benchmark through the creation of answers to the diverse questions which comprised the question template set.  Figure 8 shows an example with the question ``From 2015 to 2021, was the rate of asthma increasing faster in Boston or LA?''.  
\\
\indent
The MIMIC-III ICU QA was created using the same approach as described above. However, questions were instead based on clinical time-series data provided by MIMIC-III \cite{johnson2016mimic}. We outline all steps, along  with the code, to reproduce the benchmarks at \url{https://github.com/philipmit/thread}. We release the full dataset for DataCommons QA. Due to restrictions with MIMIC-III data access, we cannot directly release the dataset for MIMIC-III ICU QA. However, upon gaining access to MIMIC-III following the instructions outlined at \url{https://physionet.org/content/mimiciii/1.4/}, you can reproduce the full dataset using the code that we release.
 \vspace{-2mm}
\begin{wrapfigure}{r}{0.45\textwidth}
   
  \begin{center}
    \includegraphics[width=.42\textwidth]{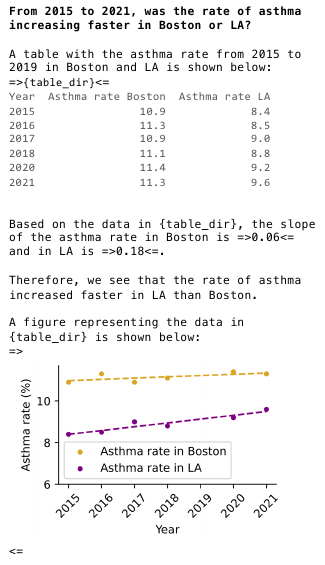}
    \caption{Question and example response with the DataCommons QA benchmark. }
  \end{center}
\end{wrapfigure}

\newpage
\onecolumn
\section{Prompts}
\label{appendix_prompts}
Below, we provide examples of the prompt used for each benchmark. The given task or question is highlighted in yellow. The text corresponding to the spawning of child threads is highlighted in blue, pink, orange, and green. The full prompts for each task can be seen at \url{https://github.com/philipmit/thread}. 

\subsection{DataCommons QA}

\rule{0.95\textwidth}{0.5pt}
\begin{small}
\raggedright
{\sethlcolor{yellow}\hl{Question: From 2018 to 2021 was the percentage of people with diabetes increasing faster in Willows or Villa Ridge?}}\\
{\sethlcolor{lightblue}\hl{Here is a table showing the percentage of people with diabetes during 2018 to 2021 for Willows and Villa Ridge: =>}}\{table\_dir\}<=\\
{\sethlcolor{lightpink}\hl{Based on the data in \{table\_dir\} from 2018 to 2021 the slope of the percentage of people with diabetes in Willows was =>}}0.3357<= and in Villa Ridge was =>0.1642<= \\
Therefore, the percentage of people with diabetes was increasing faster in Willows.\\
Final Answer: Willows \\
\#END\#
\\
\,\\
{\sethlcolor{lightblue}\hl{Here is a table showing the percentage of people with diabetes during 2018 to 2021 for Willows and Villa Ridge: =>}}\\
location1 = `Willows'\\
location2 = `Villa Ridge'\\
timeframe = `2018 to 2021'\\
variable = `percentage of people with diabetes'\\
{\sethlcolor{lightgreen}\hl{\# The \{variable\} during \{timeframe\} for location number 1, \{location1\}, is =>}}\{data\_location1\_in\_timeframe\}<=\\
\# The \{variable\} during \{timeframe\} for location number 2, \{location2\}, is =>\{data\_location2\_in\_timeframe\}<=\\
data\_location1\_in\_timeframe = [data\_location1\_in\_timeframe] if not isinstance(data\_location1\_in\_timeframe, list) else data\_location1\_in\_timeframe\\
data\_location2\_in\_timeframe = [data\_location2\_in\_timeframe] if not isinstance(data\_location2\_in\_timeframe, list) else data\_location2\_in\_timeframe\\
timeframe\_dates = np.unique([x[`date'] for x in data\_location1\_in\_timeframe] + [x[`date'] for x in data\_location2\_in\_timeframe])\\
data\_location1\_in\_timeframe\_values = ...\\
data\_location2\_in\_timeframe\_values = ...\\
...\\
...\\
table = pd.DataFrame({'date': timeframe\_dates, location1: data\_location1\_in\_timeframe\_values, location2: data\_location2\_in\_timeframe\_values})\\
table\_dir = DATA\_DIR\\
table.to\_csv(table\_dir, sep=`\t', index=False)\\
print(`\{table\_dir\}')\\
\#END\#
\\
\, \\
{\sethlcolor{lightgreen}\hl{\# The percentage of people with diabetes during 2018 to 2021 for location number 1, Willows, is =>}}\\
location1 = `Willows'\\
timeframe = `2018 to 2021'\\
time1 = `2018'\\
time2 = `2021'\\
variable = `percentage of people with diabetes'\\
query = f'\{variable\} in \{location1\}'\\
response = requests.post(URL\_EXPLORE + `?q=' + query, headers=HEADERS, json=JSON)\\
response = json.loads(response.text)\\
> print(response[`variables']) =>[`dc/topic/Diabetes', `Percent\_Person\_WithDiabetes', `Percent\_Person\_20OrMoreYears\_WithDiabetes', `dc/topic/PopulationWithDiabetesByAge', `Count\_Person\_20To79Years\_Diabetes\_AsFractionOf\_Count\_Person\_20To79Years', `WHO/SDG\_SH\_DTH\_RNCOM\_DiabetesMellitus', `dc/topic/DiabetesFemalePopulationByAge', `dc/nh3s4skee5483']<=\\
variable\_ID = `Percent\_Person\_WithDiabetes'\\
> print(response[`entities']) =>[`geoId/0685684']<=\\
location\_ID = `geoId/0685684'\\
\# I will now use the location\_ID and variable\_ID to get the data.\\
data\_location1\_all = collect\_data\_commons1(location\_ID, variable\_ID)\\
data\_location1\_in\_timeframe = [x for x in data\_location1\_all if (int(x[`date'])>=int(time1) and int(x[`date'])<=int(time2))]\\
> print(data\_location1\_in\_timeframe) =>[{'date': `2018', `value': 10.1}, {'date': `2020', `value': 10.8}, {'date': `2021', `value': 11.1}]<=\\
print(`You have retrieved \{data\_location1\_in\_timeframe\}.')\\
\#END\#\\
\, \\
\, \\
{\sethlcolor{lightpink}\hl{Based on the data in /path/to/data\_table.txt from 2018 to 2021 the slope of the percentage of people with diabetes in Willows was =>}}\\
table\_dir = `/path/to/data\_table.txt'\\
variable = `percentage of people with diabetes'\\
location = `Willows'\\
table = pd.read\_csv(table\_dir, sep=`\t')\\
\# > print(table) =>\\
\#    date  Willows  Villa Ridge\\
\# 0  2018     10.1          8.2\\
\# 1  2020     10.8          8.5\\
\# 2  2021     11.1          8.7\\
\# <=\\
location\_col\_name = `Willows'\\
X = table[`date']\\
y = table[location\_col\_name]\\
slope, intercept, r\_value, p\_value, std\_err = stats.linregress(X, y)\\
print(slope)\\
\#END\#
\end{small}
\\
\rule{0.95\textwidth}{0.5pt}

\subsection{MIMIC-III ICU QA}

\rule{0.95\textwidth}{0.5pt}
\begin{small}
\raggedright
{\sethlcolor{yellow}\hl{Question: Was the average systolic blood pressure of patient X from hour 10 to 20 of their ICU stay higher or lower than the average among all patients who expired in the hospital?}}\\
{\sethlcolor{lightblue}\hl{Here is a table showing the systolic blood pressure of patient X from hour 10 to 20 of their ICU stay: =>}}\{table\_dir\}<=\\
{\sethlcolor{lightpink}\hl{Based on the data in \{table\_dir\} the average systolic blood pressure of patient X was =>}}138.18<= \\
{\sethlcolor{lightorange}\hl{The average systolic blood pressure among all patients who expired in the hospital was =>}}120.81<= \\
Therefore, the average systolic blood pressure of patient X from hour 10 to 20 was higher than the average among all patients who expired in the hospital. \\
Final Answer: higher \\
\#END\#
\, \\
\, \\
{\sethlcolor{lightblue}\hl{Here is a table showing the systolic blood pressure of patient X from hour 10 to 20 of their ICU stay: =>}}\\
patient\_ID = `X'\\
timeframe = `hour 10 to 20'\\
variable = `systolic blood pressure'\\
{\sethlcolor{lightgreen}\hl{\# The \{variable\} during \{timeframe\} for patient \{patient\_ID\} was =>}}\{data\_patient\_in\_timeframe\}<=\\
...\\
...\\
table = pd.DataFrame({'time': timeframe\_dates, variable: data\_patient\_in\_timeframe\_values})\\
table\_dir = DATA\_DIR\\
table.to\_csv(table\_dir, sep=`\t', index=False)\\
print(`\{table\_dir\}')\\
\#END\#
\, \\
\, \\
{\sethlcolor{lightgreen}\hl{\# The systolic blood pressure during hour 10 to 20 for patient X was =>}}\\
patient\_ID = `X'\\
timeframe = `hour 10 to 20'\\
time1 = `10'\\
time2 = `20'\\
variable = `systolic blood pressure'\\
...\\
...\\
print(`You have retrieved \{data\_patient\_in\_timeframe\}.')\\
\#END\#\\
\, \\
\, \\
{\sethlcolor{lightpink}\hl{Based on the data in /path/to/data\_table.txt the average systolic blood pressure of patient X was =>}}\\
table\_dir = `/path/to/data\_table.txt'\\
variable = `systolic blood pressure'\\
table = pd.read\_csv(table\_dir, sep=`\t')\\
\# > print(table) =>\\
...\\
...\\
variable\_col\_name = `Systolic blood pressure'\\
y = table[variable\_col\_name]\\
y\_mean=np.mean(y)\\
print(y\_mean)\\
\#END\#\\
\, \\
\, \\
{\sethlcolor{lightorange}\hl{The average systolic blood pressure among all patients who expired in the hospital was =>}}\\
variable = `systolic blood pressure'\\
table = pd.read\_csv(PATIENT\_DATA\_DIR, sep=`\t')\\
\# > print(table) =>\\
...\\
...\\
filter\_col\_name = `MORTALITY\_INHOSPITAL'\\
variable\_col\_name = `Systolic blood pressure'\\
target\_patient\_group = table[table[variable\_col\_name]==1]\\
y = target\_patient\_group[variable\_col\_name]\\
y\_mean=np.mean(y)\\
print(y\_mean)\\
\#END\#
\end{small}
\\
\rule{0.95\textwidth}{0.5pt}

\subsection{ALFWorld}
\rule{0.95\textwidth}{0.5pt}
\begin{small}
\raggedright
{\sethlcolor{yellow}\hl{
\\
You are in the middle of a room. Looking quickly around you, you see a cabinet 4, a cabinet 3, a cabinet 2, a cabinet 1, a coffeemachine 1, a countertop 1, a diningtable 3, a diningtable 2, a diningtable 1, a drawer 1, a fridge 1, a garbagecan 1, a microwave 1, a sidetable 1, a sinkbasin 1, a stoveburner 4, a stoveburner 3, a stoveburner 2, a stoveburner 1, and a toaster 1.\\ 
Your task is to: clean some apple and put it in sidetable.}}\\  
target\_location\_ID = `sidetable 1'\\ 
cleaning\_location\_ID = `sinkbasin 1'\\ 
likely\_location\_IDs\_for\_obj =  [`fridge 1', `countertop 1', `cabinet 4', `cabinet 3', `cabinet 2', `cabinet 1', `diningtable 3', `diningtable 2', `diningtable 1', `sidetable 1', `sinkbasin 1', `garbagecan 1', `drawer 1', `stoveburner 4', `stoveburner 3', `stoveburner 2', `stoveburner 1']\\ 
obj = `apple'\\  
I need to perform the following steps:\\  
{\sethlcolor{lightblue}\hl{First, I need to find and take the \{obj\}. =>}}You have found and taken the \{obj\_ID\}.<=\\  
{\sethlcolor{lightpink}\hl{Next, I need to clean the \{obj\_ID\} with the \{cleaning\_location\_ID\}. =>}}You have cleaned the \{obj\_ID\} with the \{cleaning\_location\_ID\}.<=\\  
{\sethlcolor{lightorange}\hl{Finally, I need to put the \{obj\_ID\} in the \{target\_location\_ID\}. =>}}You have put the \{obj\_ID\} in the \{target\_location\_ID\}.<=\\ 
\#END\#\\ 
\, \\
\, \\
{\sethlcolor{lightblue}\hl{I need to find and take the apple. =>}}\\  
obj = `apple'\\  
My sub-task is to find and take the \{obj\}. I will check the locations in \{likely\_location\_IDs\_for\_obj\} one by one.\\  
location\_i = likely\_location\_IDs\_for\_obj[0]\\ 
{\sethlcolor{lightgreen}\hl{I need to check if there is the \{obj\} in the \{location\_i\}. =>}}There is no \{obj\} in the \{location\_i\}.<=\\  
location\_i = likely\_location\_IDs\_for\_obj[1]\\ 
...\\
...\\
I need to check if there is the \{obj\} in the \{location\_i\}. =>There is no \{obj\} in the \{location\_i\}.<=\\  
location\_i = likely\_location\_IDs\_for\_obj[2]\\ 
I need to check if there is the \{obj\} in the \{location\_i\}. =>There is no \{obj\} in the \{location\_i\}.<=\\  
location\_i = likely\_location\_IDs\_for\_obj[3]\\ 
I need to check if there is the \{obj\} in the \{location\_i\}. =>There is no \{obj\} in the \{location\_i\}.<=\\  
location\_i = likely\_location\_IDs\_for\_obj[4]\\ 
I need to check if there is the \{obj\} in the \{location\_i\}. =>There is no \{obj\} in the \{location\_i\}.<=\\  
location\_i = likely\_location\_IDs\_for\_obj[5]\\ 
I need to check if there is the \{obj\} in the \{location\_i\}. =>There is no \{obj\} in the \{location\_i\}.<=\\  
location\_i = likely\_location\_IDs\_for\_obj[10]\\ 
I need to check if there is the \{obj\} in the \{location\_i\}. =>You have found and taken the \{obj\_ID\} from the \{location\_i\}.<=\\  
print(`You have found and taken the \{obj\_ID\}.')\\ 
\#END\#\\ 
\, \\
\, \\
{\sethlcolor{lightgreen}\hl{I need to check if there is the apple in the fridge 1. =>}}\\  
location\_i = `fridge 1'\\ 
obj = `apple'\\  
My sub-task is to check if there is the \{obj\} in the \{location\_i\}.\\  
> go to \{location\_i\} =>The \{location\_i\} is closed.<=\\  
> open \{location\_i\} =>You open the \{location\_i\}. The \{location\_i\} is open. In it, you see a lettuce 2, a mug 2, a potato 2, and a tomato 1.<=\\  
obj\_IDs\_found =  [`lettuce 2', `mug 2', `potato 2', `tomato 1']\\  
obj\_number = `There is no obj\_number'\\ 
obj\_ID = `There is no obj\_ID'\\  
print(`There is no \{obj\} in the \{location\_i\}.')\\ 
\#END\#\\ 
\, \\
\, \\
{\sethlcolor{lightpink}\hl{I need to clean the apple 3 with the sinkbasin 1. =>}}\\  
cleaning\_location\_ID = `sinkbasin 1'\\ 
obj\_ID = `apple 3'\\  
My sub-task is to clean the \{obj\_ID\} with the \{cleaning\_location\_ID\}.\\  
> go to \{cleaning\_location\_ID\} =>On the \{cleaning\_location\_ID\}, you see nothing.<=\\  
> clean \{obj\_ID\} with \{cleaning\_location\_ID\} =>You clean the \{obj\_ID\} using the \{cleaning\_location\_ID\}.<=\\  
print(`You have cleaned the \{obj\_ID\} with the \{cleaning\_location\_ID\}.')\\ 
\#END\#\\ 
\, \\
\, \\
{\sethlcolor{lightorange}\hl{I need to put the apple 3 in the sidetable 1. =>}}\\  
target\_location\_ID = `sidetable 1'\\ 
obj\_ID = `apple 3'\\  
My sub-task is to put the \{obj\_ID\} in the \{target\_location\_ID\}.\\  
> go to \{target\_location\_ID\} =>On the \{target\_location\_ID\}, you see a cup 1, a lettuce 1, a peppershaker 3, a potato 1, and a saltshaker 1.<=\\  
> put \{obj\_ID\} in/on \{target\_location\_ID\} =>You put the \{obj\_ID\} in/on the \{target\_location\_ID\}.<=\\  
print(`You have put the \{obj\_ID\} in the \{target\_location\_ID\}.')\\ 
\#END\#
\end{small}
\\
\rule{0.95\textwidth}{0.5pt}

\subsection{WebShop}
\rule{0.95\textwidth}{0.5pt}
\begin{small}
\raggedright
{\sethlcolor{yellow}\hl{Instruction: i would like a 3 ounce bottle of bright citrus deodorant for sensitive skin, and price lower than 50.00 dollars }} \\ 
obj\_attributes = [`3 ounce bottle', `Bright citrus', `For sensitive skin'] \\ 
max\_price = 50.00 \\ 
obj = `deodorant' \\ 
I need to perform the following steps: \\ 
{\sethlcolor{lightblue}\hl{First, I need to retrieve search results for \{obj\} that are less than \{max\_price\} dollars with the attributes: \{obj\_attributes\}. =>}}You have retrieved \{results\_under\_max\_price\}.<= \\ 
{\sethlcolor{lightpink}\hl{Next, I need to identify the item in \{results\_under\_max\_price\} that matches the most attributes: \{obj\_attributes\}. =>}}You have identified \{item\_to\_purchase\}.<= \\ 
{\sethlcolor{lightorange}\hl{Finally, I need to purchase \{item\_to\_purchase\} with \{obj\_attributes\}. =>}}You have purchased \{item\_to\_purchase\}.<= \\ 
\#END\# \\ 
\, \\
\, \\
{\sethlcolor{lightblue}\hl{I need to retrieve search results for deodorant that are less than 50.00 dollars with the attributes: [`3 ounce bottle', `Bright citrus', `For sensitive skin']. =>}} \\ 
obj = `deodorant' \\ 
max\_price = 50.00 \\ 
obj\_attribute1 = `3 ounce bottle' \\ 
obj\_attribute2 = `Bright citrus' \\ 
obj\_attribute3 = `For sensitive skin' \\ 
> search[\{obj\}; \{obj\_attribute1\}; \{obj\_attribute2\}; \{obj\_attribute3\}] => \\ 
\text{[Back to Search]} \\ 
Page 1 (Total results: 50)  \\ 
\text{[Next >]}  \\ 
\text{[B08KBVJ4XN]}  \\ 
Barrel and Oak - Aluminum-Free Deodorant, Deodorant for Men, Essential Oil-Based Scent, 24-Hour Odor Protection, Cedar \& Patchouli Blend, Gentle on Sensitive Skin (Mountain Sage, 1 oz, 2-Pack) \\ 
\$15.95   \\ 
\text{[B078GWRC1J]}  \\ 
Bright Citrus Deodorant by Earth Mama | Pregnancy and Breastfeeding, Contains Organic Calendula 3-Ounce  \\ 
\$10.99  \\ 
\text{[B078GTKVXY]}  \\ 
Ginger Fresh Deodorant by Earth Mama | Pregnancy and Breastfeeding, Contains Organic Calendula 3-Ounce  \\ 
\$60.99 \\ 
<= \\ 
I will now filter for results that are less than \{max\_price\}. \\ 
results = [`B08KBVJ4XN', `B078GWRC1J', `B078GTKVXY'] \\ 
results\_prices = [15.95, 10.99, 60.99] \\ 
results\_under\_max\_price = [result for result, price in zip(results, results\_prices) if price < max\_price] \\ 
print(`You have retrieved \{results\_under\_max\_price\}.') \\ 
\#END\# \\ 
\, \\
\, \\
{\sethlcolor{lightpink}\hl{I need to identify the item in [`B08KBVJ4XN', `B078GWRC1J'] that matches the most attributes: [`3 ounce bottle', `Bright citrus', `For sensitive skin']. =>}} \\ 
item1 = `B08KBVJ4XN' \\ 
item2 = `B078GWRC1J' \\ 
obj\_attributes = [`3 ounce bottle', `Bright citrus', `For sensitive skin'] \\ 
My sub-task is to identify the item in [\{item1\}, \{item2\}] that matches the most attributes: \{obj\_attributes\}. \\ 
{\sethlcolor{lightgreen}\hl{I need to count the number of attributes in \{obj\_attributes\} that \{item1\} has. =>}}This item has 1 attribute.<= \\ 
I need to count the number of attributes in \{obj\_attributes\} that \{item2\} has. =>This item has 2 attributes.<= \\ 
items = [\{item1\}, \{item2\}] \\ 
item\_attributes = [1, 2] \\ 
item\_to\_purchase = next(iter(items[item\_attributes.index(max(item\_attributes))])) \\ 
print(`You have identified \{item\_to\_purchase\}.') \\ 
\#END\# \\ 
\, \\
\, \\
{\sethlcolor{lightgreen}\hl{I need to count the number of attributes in [`3 ounce bottle', `Bright citrus', `For sensitive skin'] that B08KBVJ4XN has. =>}} \\ 
item\_to\_check = `B08KBVJ4XN' \\ 
obj\_attributes = [`3 ounce bottle', `Bright citrus', `For sensitive skin'] \\ 
My sub-task is to count the number of attributes in \{obj\_attributes\} that \{item\_to\_check\} has. \\ 
> click[\{item\_to\_check\}] => \\ 
\text{[Back to Search]} \\ 
\text{[< Prev]} \\ 
Barrel and Oak - Aluminum-Free Deodorant, Deodorant for Men, Essential Oil-Based Scent, 24-Hour Odor Protection, Cedar \& Patchouli Blend, Gentle on Sensitive Skin (Mountain Sage, 1 oz, 2-Pack)  \\ 
Price: \$15.95  \\ 
Rating: N.A. \\ 
\text{[Description]} \\ 
\text{[Features]} \\ 
\text{[Reviews]} \\ 
\text{[Attributes]} \\ 
\text{[Buy Now]} \\ 
<= \\ 
I need to check if \{item\_to\_check\} has the attribute `3 ounce bottle'. \{item\_to\_check\} does not have this attribute because it is described as `1 oz, 2-Pack'. \\ 
I need to check if \{item\_to\_check\} has the attribute `Bright citrus'. \{item\_to\_check\} does not have this attribute because it is described as `Cedar \& Patchouli Blend'. \\ 
I need to check if \{item\_to\_check\} has the attribute `For sensitive skin'. \{item\_to\_check\} has this attribute because it is described as `Gentle on Sensitive Skin'. \\ 
\{item\_to\_check\} has 1 attribute. \\ 
print(`This item has 1 attribute.') \\ 
\#END\# \\ 
\, \\
\, \\
{\sethlcolor{lightorange}\hl{I need to purchase B078GWRC1J with [`3 ounce bottle', `Bright citrus', `For sensitive skin']. =>}} \\ 
item\_to\_purchase = `B078GWRC1J' \\ 
obj\_attributes = [`3 ounce bottle', `Bright citrus', `For sensitive skin'] \\ 
My sub-task is to purchase \{item\_to\_purchase\} with \{obj\_attributes\}. \\ 
> click[\{item\_to\_purchase\}] => \\ 
\text{[Back to Search]} \\ 
\text{[< Prev]} \\ 
scent [assorted scents][bright citrus][calming lavender][ginger fresh][simply non-scents] \\ 
size [travel set (4-pack)][3 ounce (pack of 1)][3-ounce (2-pack)] \\ 
Bright Citrus Deodorant by Earth Mama | Pregnancy and Breastfeeding, Contains Organic Calendula 3-Ounce  \\ 
Price: \$10.99  \\ 
Rating: N.A.  \\ 
\text{[Description]} \\ 
\text{[Features]} \\ 
\text{[Reviews]} \\ 
\text{[Attributes]} \\ 
\text{[Buy Now]} \\ 
<= \\ 
I will select from the `scent' buttons: [assorted scents], [bright citrus], [calming lavender], [ginger fresh], [simply non-scents]. \\ 
> click[bright citrus] =>You have clicked bright citrus.<= \\ 
I will select from the `size' buttons: [travel set (4-pack)], [3 ounce (pack of 1)], [3-ounce (2-pack)]. \\ 
> click[3 ounce (pack of 1)] =>You have clicked 3 ounce (pack of 1).<= \\ 
I can now select [Buy Now]. \\ 
> click[Buy Now] =>You have clicked Buy Now.<= \\ 
print(`You have purchased \{item\_to\_purchase\}.') \\ 
\#END\#
\end{small}
\\
\rule{0.95\textwidth}{0.5pt}

\subsection{TextCraft}
\rule{0.95\textwidth}{0.5pt}
\begin{small}
\raggedright
{\sethlcolor{yellow}\hl{Crafting commands: \\ 
craft 3 dark oak sign using 6 dark oak planks, 1 stick \\ 
craft 4 dark oak planks using 1 dark oak log \\ 
craft 1 stick using 1 planks \\ 
craft 4 stick using 2 bamboo \\ 
craft 4 oak planks using 1 oak log \\ 
craft 1 dark oak fence using 2 stick, 4 dark oak planks \\ 
craft 1 warped stairs using 6 warped planks \\ 
craft 3 oak sign using 6 oak planks, 1 stick \\ 
Goal: craft dark oak sign. }}\\
craft\_command\_list = ['craft 3 dark oak sign using 6 dark oak planks, 1 stick', 'craft 4 dark oak planks using 1 dark oak log', 'craft 1 stick using 1 planks', 'craft 4 stick using 2 bamboo', 'craft 4 oak planks using 1 oak log', 'craft 1 dark oak fence using 2 stick, 4 dark oak planks', 'craft 1 warped stairs using 6 warped planks', 'craft 3 oak sign using 6 oak planks, 1 stick'] \\ 
craft\_command\_item\_list = ['dark oak sign', 'dark oak planks', 'stick', 'stick', 'oak planks', 'dark oak fence', 'warped stairs', 'oak sign'] \\ 
target\_item = `dark oak sign' \\ 
target\_item\_count\_total = 1 \\ 
idx = craft\_command\_item\_list.index(min([x for x in craft\_command\_item\_list if target\_item in x], key=len)) \\ 
target\_craft\_command = craft\_command\_list[idx] \\ 
{\sethlcolor{lightblue}\hl{To craft \{target\_item\}, I need to perform the following action until I have at least \{target\_item\_count\_total\} \{target\_item\}: \{target\_craft\_command\} =>}}You have completed the action.<= \\ 
\#END\# \\ 
\, \\
\, \\
{\sethlcolor{lightblue}\hl{I need to perform the following action until I have at least 1 dark oak sign: craft 3 dark oak sign using 6 dark oak planks, 1 stick =>}} \\ 
target\_item = `dark oak sign' \\ 
target\_item\_count\_total = 1 \\ 
target\_craft\_command\_result\_count = 3 \\ 
precursor1 = `dark oak planks' \\ 
precursor1\_count\_command = 6 \\ 
precursor2 = `stick' \\ 
precursor2\_count\_command = 1 \\ 
Since I need at least 1 \{target\_item\} and each action produces 3 \{target\_item\}, The number of times I need to perform the action is ceiling of 1/3, which is 1. \\ 
target\_craft\_command\_reps = 1 \\ 
precursor1\_count\_total = precursor1\_count\_command * target\_craft\_command\_reps \\ 
precursor1\_count\_total = precursor2\_count\_command * target\_craft\_command\_reps \\ 
Next, I need to get or craft \{precursor1\}. \\ 
{\sethlcolor{lightgreen}\hl{To start, I first need to check if I can get \{precursor1\_count\_total\} \{precursor1\}. =>}}You cannot get the material.<= \\ 
Since I cannot get \{precursor1\}, I need to craft it. \\ 
idx = craft\_command\_item\_list.index(min([x for x in craft\_command\_item\_list if precursor1 in x], key=len)) \\ 
precursor1 = craft\_command\_item\_list[idx] \\ 
precursor1\_craft\_command = craft\_command\_list[idx] \\ 
To craft \{precursor1\}, I need to perform the following action until I have at least \{precursor1\_count\_total\} \{precursor1\}: \{precursor1\_craft\_command\} =>You have completed the action.<= \\ 
Next, I need to get or craft \{precursor2\}. \\ 
To start, I first need to check if I can get \{precursor2\_count\_total\} \{precursor2\}. =>You cannot get the material.<= \\ 
Since I cannot get \{precursor2\}, I need to craft it. \\ 
idx = craft\_command\_item\_list.index(min([x for x in craft\_command\_item\_list if precursor2 in x], key=len)) \\ 
precursor2 = craft\_command\_item\_list[idx] \\ 
precursor2\_craft\_command = craft\_command\_list[idx] \\ 
To craft \{precursor2\}, I need to perform the following action until I have at least \{precursor2\_count\_total\} \{precursor2\}: \{precursor2\_craft\_command\} =>You have completed the action.<= \\ 
Finally, I will perform the action 1 time. \\ 
> craft \{target\_craft\_command\_result\_count\} \{target\_item\} using \{precursor1\_count\_command\} \{precursor1\}, \{precursor2\_count\_command\} \{precursor2\} =>Crafted 3 minecraft:\{target\_item\}.<= \\ 
print(`You have completed the action.') \\ 
\#END\# \\ 
\, \\
\, \\
{\sethlcolor{lightgreen}\hl{I first need to check if I can get 6 dark oak planks =>}} \\ 
get\_item = `dark oak planks' \\ 
get\_item\_count\_total = 6 \\ 
> get \{get\_item\_count\_total\} \{get\_item\} =>Could not find \{get\_item\}.<= \\ 
I cannot get the \{get\_item\}. \\ 
print(`You cannot get the material.') \\ 
\#END\#
\end{small}
\\
\rule{0.95\textwidth}{0.5pt}


\end{document}